# Elastic LiDAR Fusion: Dense Map-Centric Continuous-Time SLAM

Chanoh Park[1,2], Peyman Moghadam[1,2], Soohwan Kim[1], Alberto Elfes[1], Clinton Fookes[2], Sridha Sridharan[2]

*Abstract*— The concept of continuous-time trajectory representation has brought increased accuracy and efficiency to multi-modal sensor fusion in modern SLAM. However, regardless of these advantages, its offline property caused by the requirement of global batch optimization is critically hindering its relevance for real-time and life-long applications. In this paper, we present a dense map-centric SLAM method based on a continuous-time trajectory to cope with this problem. The proposed system locally functions in a similar fashion to conventional Continuous-Time SLAM (CT-SLAM). However, it removes the need for global trajectory optimization by introducing map deformation. The computational complexity of the proposed approach for loop closure does not depend on the operation time, but only on the size of the space it explored before the loop closure. It is therefore more suitable for long term operation compared to the conventional CT-SLAM. Furthermore, the proposed method reduces uncertainty in the reconstructed dense map by using probabilistic surface element (surfel) fusion. We demonstrate that the proposed method produces globally consistent maps without global batch trajectory optimization, and effectively reduces LiDAR noise by surfel fusion.

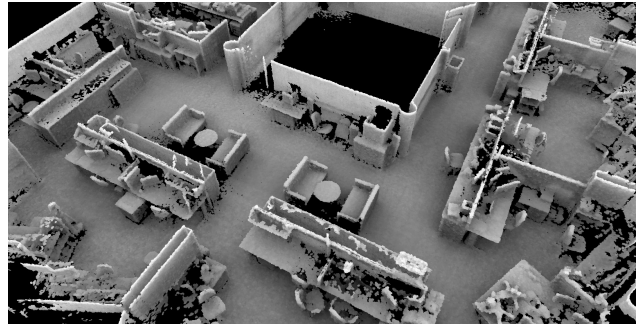

Fig. 1: Reconstructed surfel map of an office with a handheld spinning LiDAR and proposed Elastic LiDAR Fusion method. Surfels with a diameter of 20mm cover the map surface with a 10mm resolution.

## I. INTRODUCTION

In recent years, Continuous-Time Simultaneous Localization and Mapping (CT-SLAM) [1], [2] has been gaining popularity over traditional discrete-time SLAM. The main advantages of deploying CT-SLAM are twofold. Firstly, compared with conventional discrete-time SLAM approaches, CT-SLAM offers efficient approaches in combining highrate asynchronous sensors. It integrates multi-modal sensor measurements into temporally close trajectory control points, which are weighted differently to each control point according to their time of generation. In this manner, the asynchronous estimation problem is inherently addressed together. While all of the measurements are utilized in the optimization, it does not increase the dimension of the state vector. Thus, the dimensionality is decoupled from the number of measurements and is affected only by the trajectory sampling rate and the operation time without sacrificing the map accuracy.

The second main advantage of CT-SLAM is where element-wise pose recovery is required. As CT-SLAM is utilizing a continuous-time trajectory, it is relatively easy to extract a system pose at any query time. Thus, in early studies it was proven that the motion distortion problem of Light Detection and Ranging sensors (LiDAR) [3] or the rolling shutter problem of low cost image sensors [4]–[6] can be effectively solved by this property.

While high-rate asynchronous sensor fusion and element-wise pose interpolation is successfully addressed by the approach in CT-SLAM, the fact that its computational complexity for global consistency increases according to the operation time is hindering its expansion to real-time robot applications or long-term SLAM [3], [7]. The need for global batch optimization builds up the system load as time goes on and in the end it will reach an intractable level. This problem becomes critical when it comes to mobile robot applications where not only global mapping and localization but also long-term operation should be guaranteed. Furthermore, most of CT-SLAM approaches are focusing more on trajectory optimization and therefore, they are not able to take advantage of multiple observations of a space for better dense mapping.

This paper proposes a new approach for dense LiDAR mapping that combines conventional CT-SLAM with a map-centric approach inspired by [8] to overcome limitations mentioned above. The proposed approach operates in a similar fashion to local CT-SLAM. However, to remove the global trajectory dependency of the loop closure we build up the global map without a global trajectory and deform the space itself for a loop closure. As the proposed method does not maintain a trajectory, we propose a map prior constraint for the local trajectory optimization so the map itself can play the role of a trajectory. Furthermore, a dense global map is built based on map element-wise probabilistic fusion in a way that all of the spatially redundant LiDAR measurements are fully utilized and merged into the canonical form of the space. Therefore, the proposed method has a strong advantage in long-term operation over conventional trajectory-centric CT-SLAM.

[1] The authors are with the Autonomous Systems, Cyber-Physical Systems, DATA61, CSIRO, Brisbane, QLD 4069, Australia. E-mails: *firstname.lastname*@data61.csiro.au
[2] The authors are with the School of Electrical Engineering and Computer Science, Queensland University of Technology (QUT), Brisbane, Australia. E-mails: *chanoh.park, peyman.moghadam, c.fookes, s.sridharan*@qut.edu.au

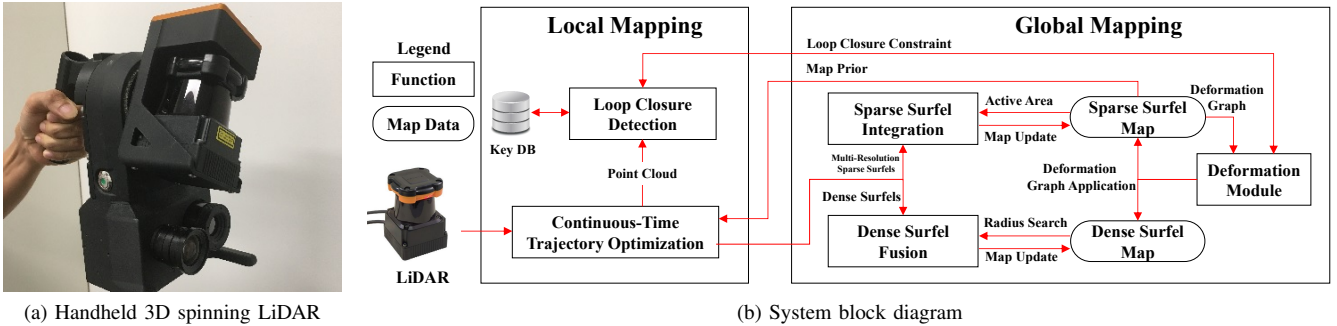

(a) Handheld 3D spinning LiDAR  (b) System block diagram

Fig. 2: (a) The experimental handheld 3D spinning LiDAR for mobile mapping. It contains a 2D laser, an IMU, an encoder, a color camera and a thermal camera. Note that the thermal camera is not used in the paper. (b) System block diagram of our method. The device local trajectory is tracked in the Local Mapping stage, while the global consistent map is maintained in the second Global Mapping stage.

This paper is organized as follows. Section 2 reviews related works. In Section 3, we describe the overall architecture of our system, and detail our approach in Section 4, 5 and 6. Results demonstrating advantages of our method over conventional CT-SLAM are given in Section 7. Finally we summarize and present future directions in Section 8.

## II. RELATED WORK

In 2009, Bosse and Zlot [9] proposed a linear interpolation based continuous-time scan matching with a spinning 2D LiDAR. Their method subsampled vehicle odometer readings and applied geometric feature matching constraints on the related two trajectory points. The suggested linear continuous-time trajectory addressed the motion distortion problem, however, as it was an open loop approach, global map consistency was not guaranteed. In their later research [3], they further extend their work to a complete SLAM system by introducing a global trajectory optimization. However, the introduction of the global batch trajectory optimization required an offline processing step once scanning is done.

More recent researches has extended the approach in [3] by loosely coupled Visual-LiDAR constraint [10], tightly coupled Visual-inertial constraint [7], [5], and tightly coupled Visual-Inertial-LiDAR constraint [2], yet all of these approaches suffer from the same problem.

Although the continuous-time trajectory approach successfully addressed multi-modal sensor fusion [11] and motion distortion problems, their applications are limited to offline-based processing, because of the necessity of the entire trajectory and increasing computational complexity. Thus, it is not applicable to real-time robotic applications where long-term operation should be guaranteed as well as global consistency.

Recently, ElasticFusion [8] proposed a map-centric approach that removes pose graph optimization yet performs globally consistent mapping by deforming the map itself. The concept of map-centric SLAM eliminated the need of a pose graph for globally consistent mapping and converted the time dependency of the global optimization to a space-dependent problem. Also, by confining the tracking and fusion within recent map elements, they drastically reduced the processing time per input frame. However, some features of their approach, such as projective data association and confidence based surfel fusion, are limited to RGB-D sensors and are not applicable to LiDAR sensor model [12].

Thus, in this paper we adopt the map-centric approach in [8] to improve applicability of conventional CT-SLAM in the lifelong applications.

## III. OVERVIEW

Our proposed CT-SLAM system is composed of two main components: local mapping and global mapping as shown in Figure 2. The first stage of the system takes Inertial Measurement Unit (IMU) and LiDAR measurements to build motion-distortion-corrected maps by continuous-time trajectory optimization. This stage is similar to the sliding window operation in CT-SLAM [1], [3], [9], but the main difference is that it takes the sparse global map as a map prior for localization. The loop closure detection module keeps generating key points of the geometry in the moving window and compares them to the previously generated key frames.

The second stage of the proposed system builds and fuses surfel maps. Note that the two different maps, multi-resolutional ellipsoid surfel map and disk surfel map, which are denoted as sparse and dense maps, are utilized for different purposes. The multi-resolutional 3D ellipsoidal surfels in the sparse map proposed in [9] are ideal for fast and robust continuous-time trajectory optimization, whereas it is too sparse for dense mapping [13]. Thus, we utilize 2D disk surfels [8], [14] for dense surfel fusion.

Global map consistency is achieved by non-rigid deformation using the loop closure constraint from the loop closure module. This part of our method is inspired by the global deformation part of ElasticFusion [8]. We also extend it by considering surfel uncertainty propagation. Upon loop closure detection, non-rigid deformation is performed on a graph to reduce the error of loop closure constraints. After the optimization of the deformation graph, the deformation is applied to the entire map.

## IV. LOCAL MAPPING

When a LiDAR sensor is capturing continuously on a moving platform, LiDAR scans become locally motion distorted. In this section, we describe how the geometrical and inertial information from measurements can be utilized to compensate for such motion distortion, combined with the continuous-time trajectory optimization. At the end of this stage, a locally consistent point cloud will be generated according to the corrected trajectory. As the proposed method directly utilizes the global map prior as a constraint, the reprojected point cloud is always represented in the global map coordinates.

### A. Continuous-Time Trajectory Representation

Before presenting the detailed description of the local mapping, we will review the concept and advantages of the continuous-time trajectory representation.

The first important property of the continuous-time trajectory representation is that the trajectory is modeled as a function of time. An exact system pose $\mathbf{T}_\tau \in SE(3)$ at an arbitrary query time $\tau \in \mathbb{R}$ can be interpolated from a set of trajectory elements $\mathbb{Q}$. A single discrete trajectory element $\mathbf{Q}_k \in \mathbb{Q}$ is created at an interval of $\tau_{step}$ with an initial prior. Let $\mathbf{T}_\tau$ be composed of translational component $\mathbf{t}_\tau \in \mathbb{R}^3$ and rotational component $\mathbf{R}_\tau \in SO(3)$,

$$\mathbf{T}_\tau = \begin{bmatrix} \mathbf{R}_\tau & \mathbf{t}_\tau \\ \mathbf{0} & 1 \end{bmatrix} \quad (1)$$

Then, $\mathbf{T}_\tau$ is defined as an interpolation of neighboring trajectory elements $\mathbf{Q}_{k \in \Phi(\tau)}$ where $\Phi(\tau)$ represents a set of neighbor indices. There are two types of well-known interpolation techniques for the continuous trajectory. The trajectory element $\mathbf{Q}_k$ will be a direct pose $\mathbf{T}_k \in SE(3)$, $\mathbf{Q}_{k \in \Phi(\tau)} = \{\mathbf{R}_k, \mathbf{t}_k, \mathbf{R}_{k+1}, \mathbf{t}_{k+1}\}$ for the linear case [9] or indirect control points $\mathbf{c}_{\mathbf{r}k}, \mathbf{c}_{\mathbf{t}k} \in \mathbb{R}^3$, $\mathbf{Q}_{k \in \Phi(\tau)} = \{\mathbf{c}_{\mathbf{t}k-1}, \mathbf{c}_{\mathbf{r}k-1}, \cdots \mathbf{c}_{\mathbf{t}k+2}, \mathbf{c}_{\mathbf{r}k+2}\}$ for the B-spline case [5], [7].

Although, the B-spline trajectory representation is $C^2$ continuous and offers easy formulation for a derivative of the trajectory, its motion resolution and computational cost is inversely proportional. Hence, it has been applied for the offline applications [7] or used in real-time application with a low frequency motion assumption [15].

In contrast, the linear case leaves the raw high-frequency motion and only adjusts the low-frequency part of the trajectory by optimizing the subsampled trajectory and applying the low-frequency compensation to the original trajectory. In such way, the trajectory can represent a higher-frequency motion whereas the computational cost is much lower than the B-spline case. For this reason, we opted for the linear case.

In the linear interpolation case $\mathbf{T}_\tau$ is defined from two poses where their timestamps satisfy $\tau_k < \tau < \tau_{k+1}$. Thus, given poses, their interpolation at $\tau$ is given by,

$$\mathbf{R}_\tau = \mathbf{R}_k \mathbf{e}^{\alpha[\omega]_\times}, \quad \mathbf{t}_\tau = (1-\alpha)\mathbf{t}_k + \alpha \mathbf{t}_{k+1} \quad (2)$$

where the relative motion $\omega \in \mathbb{R}^3$ of the exponential mapping $\mathbf{e}^{[\omega]_\times}$, that linearly interpolates the rotational component of two positions on the manifold, is defined as $\omega = ln(\mathbf{R}_k^{-1}\mathbf{R}_{k+1})$ and the interpolation ratio $\alpha = (\tau - \tau_k)/(\tau_{k+1} - \tau_k)$.

In addition to the first property; ability of evaluating a pose at arbitrary time $\tau$, the second advantage is the efficiency in applying constraints to the trajectory elements under the circumstances where we need to carefully deal with asynchronous and high-rate sensors. Let $f_n \in \mathbb{R}$ be an objective function defined for each constraint $n$ which takes sensor measurements $\mathbf{z}_n$ and a system pose $\mathbf{T}_\tau$ as inputs. Then, our objective is to find the corrected trajectory elements $\hat{\mathbb{Q}}$ that minimize the error functions.

$$\hat{\mathbb{Q}} = \underset{\mathbb{Q}}{\operatorname{argmin}} \sum_n \sum_l f_n(\mathbf{T}_\tau(\mathbf{Q}_{k \in \Phi(\tau_\mathbf{z})}), \mathbf{z}_{n,l}) \quad (3)$$

where $l$ represents the $l$-th measurement from the $n$-th constraint and $\tau_\mathbf{z}$ represents the timestamp of each measurement. Equation (3) can be solved by an iterative nonlinear least square method.

As $\mathbf{T}_\tau$ are defined as an interpolation of trajectory elements, high rate sensor measurements increase the number of rows in its Jacobian instead of the number of the state dimensions to be optimized. This makes an obvious difference from the conventional discrete-time trajectory model, where poses are created whenever there is a new sensor observation. An example of different objective functions $f_n$ will be described in the following section.

### B. Deformation Handling by Local Trajectory Optimization

When the motion of a device is relatively moderate compared to the scanning speed and the initial trajectory guess is fairly good, shapes of local features are often well preserved in the measurements even with the motion distortion. The first two geometrical constraints utilize this property of sweeping type LiDAR measurements [9], [10]. This process starts by transforming the LiDAR measurements with respect to the world frame and extracting sparse ellipsoidal surfels from multi-resolutional voxels [9]. From the sparse surfels, we calculate the distance errors between two corresponding new surfels $a$ and $b$ and between a new surfel $c$ and its corresponding global map surfel $m$ in their averaged normal directions as illustrated in Figure 3.

$$\mathbf{e}_I = \sum \|\mathbf{n}_{ab}^T(\mathbf{R}_{\tau_a}\mathbf{u}_{\tau_a} + \mathbf{t}_{\tau_a} - (\mathbf{R}_{\tau_b}\mathbf{u}_{\tau_b} + \mathbf{t}_{\tau_b}))\|^2 \quad (4)$$

$$\mathbf{e}_M = \sum \|\mathbf{n}_{mc}^T(\mathbf{u}^m - (\mathbf{R}_{\tau_c}\mathbf{u}_{\tau_c} + \mathbf{t}_{\tau_c}))\|^2 \quad (5)$$

where $\mathbf{u}_\tau$ is the centroid of a surfel, $\mathbf{t}_\tau, \mathbf{R}_\tau$ are the interpolated sensor pose at time $\tau$. Note that the map prior $\mathbf{u}^m$ does not need a transformation, as they are points in the world coordinate system. The motion distortion is corrected based on the map prior. The initial map prior is given from a short period of stationary scanning in the beginning.

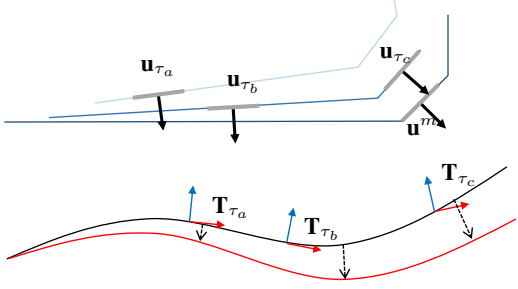

Fig. 3: Illustration of geometrical constraints on the local trajectory. Two surfels $\mathbf{u}_{\tau_a}$ and $\mathbf{u}_{\tau_b}$ generated at $\tau_a$, $\tau_b$ within a local window forms constraints on the interpolated trajectory poses $\mathbf{T}_{\tau_a}$ and $\mathbf{T}_{\tau_b}$. The constraint between the local sweep $\mathbf{u}_{\tau_c}$ and the map prior $\mathbf{u}^m$ forces the trajectory to be fitted onto the map prior.

On the other hand, the inertial information from IMU offers a prior regarding rotational velocity and translational acceleration of the trajectory. As the observed rotational velocity and translational acceleration from IMU should coincide with that of the trajectory, the constraints are given as,

$$\mathbf{e}_\alpha = \sum \|\boldsymbol{\alpha}_\tau - \mathbf{R}_\tau^T(\frac{d^2}{d\tau^2}\mathbf{t}_\tau - \mathbf{g}) + \mathbf{b}_\alpha\|^2 \quad (6)$$

$$\mathbf{e}_\omega = \sum \|\boldsymbol{\varpi}_\tau - \boldsymbol{\omega}_\tau + \mathbf{b}_\omega\|^2 \quad (7)$$

where $\boldsymbol{\alpha}$, $\boldsymbol{\varpi}$ are respectively acceleration and rotational velocity from IMU measured at time $\tau$, $\mathbf{g}$ is the gravitational acceleration, $\mathbf{R}_\tau, \mathbf{t}_\tau, \boldsymbol{\omega}_\tau$ are respectively interpolated rotation, translation, and rotational velocity of the trajectory at $\tau$, and $\mathbf{b}_\alpha, \mathbf{b}_\omega$ are the IMU biases. These two constraints form a strong restriction on local smoothness but they are prone to error in a relatively long period, *e.g.*, a few seconds, as they are first and second derivative constraint. Geometrical constraints of Equation (4), (5) address this problem and help optimization to be locally and globally consistent.

Finally, the corrected trajectory is achieved by finding a set of parameters that minimize the constraints above as,

$$\hat{x} = \underset{x}{\operatorname{argmin}} \; \mathbf{e}_I + \mathbf{e}_M + \mathbf{e}_\omega + \mathbf{e}_\alpha \quad (8)$$

where $x = [\mathbb{Q}, \mathbf{b}_\omega, \mathbf{b}_\alpha, d]$.

After the optimization, local dense and sparse surfel maps $\mathbb{M}_l, \mathbb{S}_l$ are built and updated by the optimized trajectory. Each 2D disk surfel in the local dense map is composed of position $\mathbf{p} \in \mathbb{R}^3$, surfel normal $\mathbf{n} \in \mathbb{R}^3$, and timestamp $t$. Also, the uncertainties of position and normal $\Sigma_\mathbf{p} \in \mathbb{R}^{3\times 3}, \Sigma_\mathbf{n} \in \mathbb{R}^{3\times 3}$, which are utilized in dense surfel matching and fusion are calculated from their neighboring points. On the other hand, local sparse surfel maps $\mathbb{S}_l$ which are utilized in Equation (4), (5) are updated by the optimized trajectory. Each 3D ellipsoid surfel is defined with a centroid $\mathbf{c} \in \mathbb{R}^3$ and a covariance matrix $\Sigma_\mathbf{c} \in \mathbb{R}^{3\times 3}$ which represents the distribution of points within the voxel. The local maps $\mathbb{M}_l, \mathbb{S}_l$ are fused into the global maps $\mathbb{M}_g, \mathbb{S}_g$ in the following section.

## V. SURFEL FUSION

In this section, temporal and probabilistic fusion to fully utilize multiple observations of a scene will be discussed.

### A. Surfel Matching and Fusion

We modified the fusion method for the dual surfel map from our previous work [12]. As the property of each sparse and dense surfel map $\mathbb{M}_g, \mathbb{S}_g$ is different, different approaches for data association and fusion should be taken to each map. For sparse surfel map $\mathbb{S}_g$, considering that planarity of a surfel is important in continuous-time trajectory optimization, the surfel map is updated upon finding a wider and thinner surfel within a sparse map resolution. The dense surfel $\mathbb{M}_g$, on the other hand, where the utilized number of points is much smaller than the sparse surfel and as such is prone to sensor noise, requires a dedicated approach for data association and fusion. For data association, it utilizes a sensor noise model to search its match deeper along the beam direction, while also densifying map resolution by introducing the resolutional threshold. To effectively handle sensor noise, surfels are fused based on Bayesian fusion that considers the sensor noise model.

For the colorization of the dense mapping, we utilized confidence based color fusion proposed in [16] with additional parameters to consider uncertainties caused by fisheye lens distortion and depth.

### B. Active and Inactive Maps

One of the most important assumptions in the fusion is that the global map consistency around the area where a fusion occurs should be guaranteed. However, as we are deploying an on-the-fly loop closure method, this is not the case. In the worst case, surfels will be fused right before the loop closure and will entirely ruin the local area as in Figure 4 (b). Thus, similar to [8], we introduce active and inactive maps $\mathbb{A}, \mathbb{I}$ according to surfel timestamps so that new surfels are always optimized and fused within the active area. To detect the misalignment between the active area and the inactive area, it finds the amount of overlap and misalignment by Iterative Closest Point (ICP). For a robust and fast ICP, only sparse surfels are utilized with a geometrical weight as in [12]. When the overlap and misalignment is large enough after the ICP, it triggers deformation that will be described in the next section. To maintain the map coherency between active and inactive components, matched inactive components to active components are found and updated in every step. Figure 4 and Algorithm 1 show our temporal fusion method.

## VI. GLOBAL MAP BUILDING

As the proposed system does not maintain any trajectory, an alternative method for a loop closure is necessary. In this section, we describe our deformation-based approach for maintaining the map to be globally consistent.

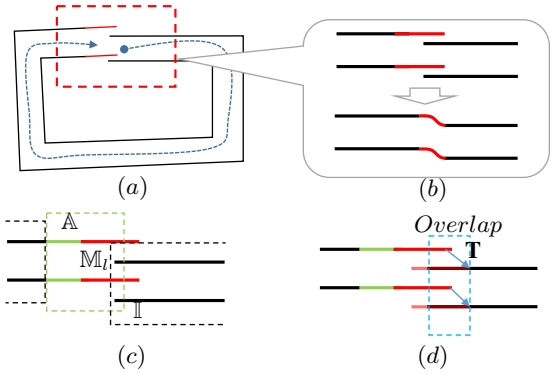

Fig. 4: (a) A map including a loop closure. Red solid lines represent the latest map elements. (b) Fusion at the loop. Before the fusion (upper) and after the fusion (lower) (c) Fusion only within an active window (solid green lines). (d) Misalignment detection between inactive (solid black lines) and active area.

**Algorithm 1:** Temporal Fusion. To prevent the fusion right before loop closure, the proposed method divides the map into active and inactive regions according to the timestamps of each surfel and fuses within active areas only. A misalignment between active and inactive is detected by ICP.

**Input:** New Frame $\mathbb{S}_l, \mathbb{M}_l$, Global Maps $\mathbb{S}_g, \mathbb{M}_g$
**Output:** Updated Global Maps $\mathbb{S}_g, \mathbb{M}_g$

1 $\mathbb{A} \leftarrow GetCurrentActive(\mathbb{S}_g, \mathbb{M}_g)$
2 $\mathbb{I} \leftarrow GetCurrentInactive(\mathbb{S}_g, \mathbb{M}_g)$
3 $\mathbb{S}_g, \mathbb{M}_g \leftarrow Fusion(\mathbb{S}_l, \mathbb{M}_l, \mathbb{A})$
4 $[\mathbf{R}, \mathbf{t}, inlier, dist] \leftarrow ICP(\mathbb{S}_l, \mathbb{I})$
5 **if** $inlier > \theta_{in}$ & $dist > \theta_d$ **then**
6 $\quad [\hat{\mathbf{R}}_j, \hat{\mathbf{t}}_j] \leftarrow GraphOptimization(\mathbf{R}, \mathbf{t}, \mathbb{S}_l, \mathbb{S}_g)$
7 $\quad \mathbb{S}_g, \mathbb{M}_g \leftarrow MapDeform(\hat{\mathbf{R}}_j, \hat{\mathbf{t}}_j, \mathbb{S}_g, \mathbb{M}_g)$
8 **else**
9 $\quad n \leftarrow NearistNeighborSearch(\mathbb{A}, \mathbb{I})$
10 $\quad$ **if** $n < \theta_n$ **then**
11 $\quad\quad \mathbb{S}_g \leftarrow Fusion(\mathbb{A}, \mathbb{I})$
12 $\quad$ **end**
13 **end**

### A. Map Deformation

Upon loop closure detection, map deformation is carried out to maintain global map consistency. Deformation nodes and loop closure constraints are only selected from the sparse surfel map $\mathbb{S}_g$. Then, once optimal deformation is found by an optimization, the deformation is applied to both entire map elements in $\mathbb{M}_g, \mathbb{S}_g$. We adopted the graph deformation technique and temporally connected deformation graph of [8]. However, our approach is different from [8] in that the number of deformation nodes is decided by the $m^2$ (square metre) area size of the reconstructed space and the uncertainty is deformed along with the normal and centroid.

*1) Graph Construction:* Graph nodes are constructed from randomly selected surfel centroids of $\mathbb{S}_g$. The centroids are selected to uniformly represent the space. The number of nodes is decided by $m^2$ surface area of the space. As the proposed dense surfel map fusion maintains a canonical form of surface without surficial redundancy, the $m^2$ size of the map surface can be easily found by counting the number of total surfels. Thus, considering that each surfel of the map has a fixed radius circle, the total number of nodes $n$ is,

$$n = \lceil \gamma \pi r^2 (E/m^2) \rceil \quad (9)$$

where $\gamma$ is the total number of map elements, $r$ is the radius of the circle, and $E$ is the number of nodes per $m^2$. Denser nodes help to reduce local spatial distortion but exponentially increase the computational cost. To reject any temporally uncorrelated connection, we order the nodes by temporal sequence and define their neighbors according to their temporal closeness.

The deformation graph is composed of node positions $\mathbf{g}_j \in \mathbb{R}^3$, node rotations and translations, $\mathbf{R}_j, \mathbf{t}_j \in SE(3)$, and a set of neighbors $\mathbb{V}(\mathbf{g}_j)$. Most of the time, the map is deformed by the difference between node translations. Regarding the number of neighbors, previous research [8] indicates that four neighbor nodes are sufficient.

*2) Graph Application:* Assuming that a set of graph node locations $\mathbf{g}_j$ are established and their deformation attributes $\mathbf{R}_j, \mathbf{t}_j$ are found, the graph deformation can be applied to the entire map. Given a set of node parameters, an influence function $\phi(\mathbf{p}_i)$ that deforms any given point $\mathbf{p}_i$ by the $j$-th node $\mathbf{g}_j$ is defined as

$$\phi(\mathbf{p}_i) = \mathbf{R}_j(\mathbf{p}_i - \mathbf{g}_j) + \mathbf{g}_j + \mathbf{t}_j \quad (10)$$

For a smooth blending of deformation, the complete deformation of a point is defined as a weighted sum of the influence function $\phi(\mathbf{p}_i)$ with its neighbor nodes $\mathbb{N}(\mathbf{p}_i)$ around the position. When selecting nodes, their temporal and spatial closeness are considered. Thus, the deformed pose $\mathbf{p}'_i$ is defined as,

$$\mathbf{p}'_i = \sum_{j \in \mathbb{N}(\mathbf{p}_i)} w_j(\mathbf{p}_i) \phi(\mathbf{p}_i) \quad (11)$$

where the weight $w_j(\mathbf{p}_i)$ is decided by the distance between the node $\mathbf{g}_j$ and the point $\mathbf{p}_i$,

$$w_j(\mathbf{p}_i) = (1 - \|\mathbf{p}_i - \mathbf{g}_j\|/d_{max}) \quad (12)$$

where $d_{max}$ represents the max distance between the node and point within $\mathbb{N}(\mathbf{p}_i)$. The nodes that are relatively far from the given point $\mathbf{p}_i$ have smaller weights and less effect. Note that Equation (11) will be applied to both sparse and dense maps $\mathbb{S}_g, \mathbb{M}_g$.

Not only surfel centroids, but other surfel attributes also need to be deformed. The new normal direction of the surfel is defined as

$$\mathbf{n}'_i = \sum_{j \in \mathbb{N}(\mathbf{p}_i)}^{k} w_j(\mathbf{p}_i) \mathbf{R}_j^{-T} \mathbf{n}_i \quad (13)$$

The uncertainty and geometry matrices $\Sigma_{\mathbf{p}i}, \Sigma_{\mathbf{n}i}, \Sigma_{\mathbf{c}i}$ are deformed by a first order linear propagation. The general form of the propagated uncertainty and geometry in the new deformed space is given as

$$\Sigma'_i = \mathbf{R}'_i \Sigma_i \mathbf{R}'^T_i \quad (14)$$

where $\mathbf{R}'_i$ represents the blended rotation of the surfel location and defined as

$$\mathbf{R}'_i = \sum_{j \in \mathbb{N}(\mathbf{p}_i)}^{k} w_j(\mathbf{p}_i) \mathbf{R}_j \quad (15)$$

*3) Graph Optimization:* In this section, we will describe constraints for the graph optimization, given a set of matched surfel centroid pairs $\mathbf{p}_{src}, \mathbf{p}_{dest} \in \mathbb{S}_g$. During the optimization, the transformation $\mathbf{R}_j, \mathbf{t}_j$ of each node which minimize the source and destination will be found. The first constraint that reduces the difference between the deformed source sets $\mathbf{p}'_{src} = \phi(\mathbf{p}_{src})$ and target surfel $\mathbf{p}_{dest}$ is

$$\mathbf{e}_{loop} = \sum \|\mathbf{p}'_{src} - \mathbf{p}_{dest}\|^2 \quad (16)$$

The distortion is applied over the all global map surfels including the destination surfels $\mathbf{p}_{dest}$ itself. To prevent the infinite loop of deformation between source and destination, the following pinning constraint will fix the destination surfels not to be deformed.

$$\mathbf{e}_{pin} = \sum \|\mathbf{p}'_{dest} - \mathbf{p}_{dest}\|^2 \quad (17)$$

To guarantee the smooth deformation over the whole region, the following term spreads the deformation to the neighboring nodes. When the node rotation $\mathbf{R}_j$ is near the identity matrix, the smoothness term indicates the distance between $\mathbf{t}_j$ and its neighboring node translation $\mathbf{t}_k$.

$$\mathbf{e}_{reg} = \sum_j^m \sum_{k \in \mathbb{V}(\mathbf{g}_j)} \|\mathbf{R}_j(\mathbf{g}_k - \mathbf{g}_j) + \mathbf{g}_j + \mathbf{t}_j - \mathbf{g}_k - \mathbf{t}_k\|^2 \quad (18)$$

Then, the optimal node transformation $\hat{\mathbf{R}}_j, \hat{\mathbf{t}}_j$ that minimizes radical deformation and loop closing error is defined as

$$[\hat{\mathbf{R}}_j, \hat{\mathbf{t}}_j] = \underset{\mathbf{R}_j, \mathbf{t}_j \in SE(3)}{\arg\min} \omega_{reg} \mathbf{e}_{reg} + \omega_{pin} \mathbf{e}_{pin} + \omega_{loop} \mathbf{e}_{loop} \quad (19)$$

where $\omega_{reg}, \omega_{pin}, \omega_{loop}$ represent the weights for each constraint which follow the proposed values in [8].

Since the cost function is a non-linear pose graph optimization problem in the form of $f(\mathbf{T}) = \mathbf{T}\mathbf{p} + \mathbf{c}$ on a manifold, we use the non-linear iterative Gauss-Newton method.

*4) Global Loop Closure:* There are two sources of loop closure detection. For a loop where the source and destination distance is moderately close, the ICP in Algorithm 1 will detect the misalignment first. However, for a large-scale mapping where a possible large extent of misalignment exists, an alternative 3D point cloud-based place recognition such as [17], [18] is required. We utilize the 3D point cloud based place recognition presented in [17] where descriptors of the input scene are calculated in every frame and compared to the keys of the scenes in the database. In both cases, loop closure constraints are given as follows.

$$\mathbf{p}_{dest} = \mathbf{R}\mathbf{p}_{src} + \mathbf{t} \quad (20)$$

where $\mathbf{p}_{src}$ are randomly selected points from $\mathbb{S}_l$ and the misalignment $\mathbf{R}, \mathbf{t}$, which is achieved from ICP between $\mathbb{S}_l$ and $\mathbb{I}$. Note that we utilize a transformed destination $\mathbf{p}_{dest}$ from the $\mathbf{p}_{src}$ to prevent any unwanted deformation caused by source and destination point difference. This way, the deformation implicitly reduces the misalignment $[\mathbf{R}, \mathbf{t}]$ to be $[\mathbf{I}, \mathbf{0}]$.

| Methods | Optimization* | No. State | Elapsed Time (sec) |
|---|---|---|---|
| Proposed | Fig. 5 (i) | 192 | 0.12 |
| CT-SLAM [3] | Fig. 5 (ii) | 3396 | 195.40 |

TABLE I: Global optimization cost comparison of the map in Figure 5. Note that the state dimension for each state is 6 DoF for both [3] and the proposed work. In the proposed work, the state vector represents the node transformation whereas in [3] it is trajectory elements. CT-SLAM optimizes the subsampled global trajectory instead of the full trajectory. *Location of loop closure occurrence.

## VII. Experiment

We present the quantitative and qualitative performance analysis of the proposed method in terms of global loop closure cost, trajectory estimation and surface estimation performance. The evaluation is made based on four different datasets which include various ranges, from a small size room to a large size indoor and outdoor mixed environment.

A hand-held 3D spinning LiDAR is utilized for the real data experiments. The device consists of a spinning Hokuyo UTM-30LX laser, an encoder, a Microstrain 3DM-GX3 IMU, Grasshopper3 2.8 MP color camera and Optris PI 450 thermal-infrared 382×288 pixel camera (Figure 2 (a)).

### A. Global Loop Closure Cost Estimation

To quantitatively compare the global optimization cost of the proposed method over the batch optimization based CT-SLAM, we utilized a scanning scenario that includes a loop closure and a redundant scanning path as shown in Figure 5 (a). The loop is closed at Figure 5 (i) in the proposed method which is right after the detection of a loop closure. After the first loop closure it goes without further loop closure until the end. However, CT-SLAM [3] does the loop closure at the end (Figure 5 (ii)) on its entire trajectory for the global batch optimization. Table I shows the cost comparison between CT-SLAM and the proposed method. The state dimension to be optimized is much smaller in the proposed method. The qualitative comparison between the trajectory optimized by global batch trajectory optimization and deformation is shown in Figure 6.

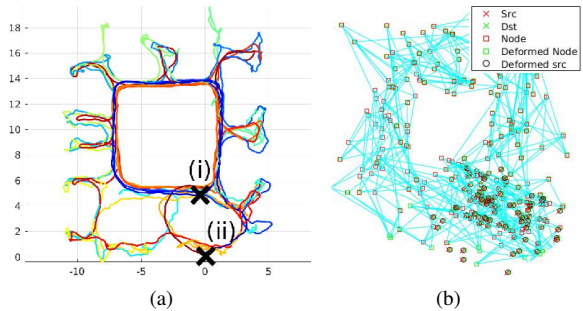

Fig. 5: Scanning trajectory and deformation graph of Figure. 1. (a) Scanning path. Our proposed method closes a loop at (i) whereas CT-SLAM does at (ii). The trajectory is colored by time: blue at the start transitioning to red at the end. (b) Constructed graph (red squares) and its correction (green squares).

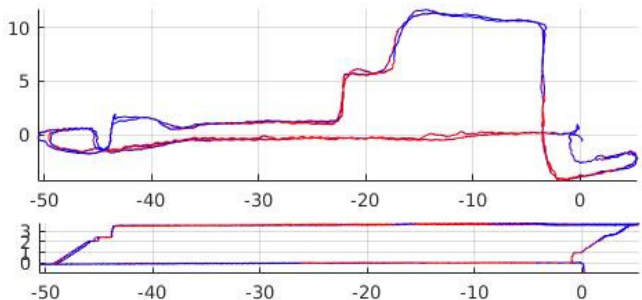

Fig. 6: Qualitative trajectory comparison between the global trajectory optimization (blue line) [3] and the proposed method (red line) of the map in Figure 7 (ii). (Upper) Top view. (Lower) Side view. Note that the trajectory includes two traverses.

### B. Trajectory Estimation

To evaluate the trajectory estimation accuracy of the proposed method, we compared the estimated trajectory with the globally optimized trajectory from [3]. For the comparison, we utilized the absolute trajectory Root Mean Square Error (RMSE) metric [8], which estimates the Euclidean distances between the distortion-based trajectory and the globally optimized trajectory. Table II shows the error and trajectory statistics of each map. A qualitative comparison of trajectories is presented in Figure 6. Note that the proposed method is not required to store the trajectory. They are saved and deformed only for a comparison.

### C. Surface Estimation

For a quantitative comparison where ground truth is not available, a well-known structure such as a floor or wall is utilized for calculating the relative noise level. Table III shows statistical analysis of the multiple planner areas (red circles in Figure 8 (b)). The error in the table represents the mean projective distance, which is calculated from the mean plane of each patch. Statistics shows that the estimated surface is up to three times less noisy than the point cloud generated using CT-SLAM [3]. Also, regardless of the number of the original raw points in the proposed method the surfel density in patches are uniformly maintained after the fusions. Additionally, with the surfel-based map representation the proposed method is capable of producing a dense 2D scenes even with sparser point cloud as shown in the last row of Figure 7. Figure 8 (a) shows the qualitative comparison of an extracted circular patch of [3] and the proposed method. Most of the noisy points in the original CT-SLAM point cloud (Figure 8 (a) left) are removed and fused into the minimum number of surfels that are required to maintain the map surface with the specified surface resolution. See our supplementary video to clearly visualize and understand the proposed method[1].

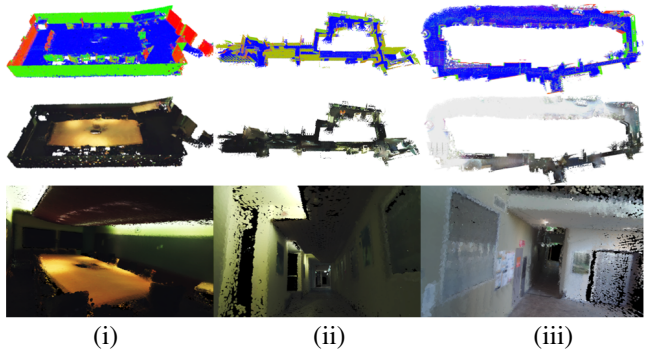

Fig. 7: Qualitative datasets. Starting from the top row, normal map, colorized map, details of the map; (i) A small size meeting room; (ii) Multiple floor structure. Map dimension and trajectory length is listed in the Table II; (iii) Indoor and outdoor mixed environment.

| Dataset | Traj Error (m) | Length (m) | Time (min) | Size (m) |
|---|---|---|---|---|
| Fig. 1 | 0.047 | 330 | 14.6 | 20×20 |
| Fig. 7 (i) | 0.041 | 130 | 6.1 | 10×6 |
| Fig. 7 (ii) | 0.056 | 300 | 11.4 | 55×20 |
| Fig. 7 (iii) | 0.076 | 360 | 9.1 | 60×25 |

TABLE II: Trajectory estimation RMSE between the deformed trajectory and the globally optimized trajectory (CT-SLAM [3]).

## VIII. CONCLUSION

In this paper, a new approach for dense LiDAR-based map-centric CT-SLAM was presented. The proposed system utilizes map deformation as a way for maintaining global map consistency instead of conventional global batch trajectory optimization to improve the applicability of the conventional CT-SLAM in long-term operation applications. As a result, our system does not need to wait until the end of scanning to achieve a globally consistent map. Also, under the guarantee of global map consistency at all times, we were able to utilize an on-the-fly map fusion method in which all the measurements are fused into the global map. Therefore, measurements storage for global optimization is not necessary. Our experimental results indicate the deformed trajectory is marginally different to the trajectory produced by the global batch optimization, however surface estimation is up to three times less noisy than the conventional CT-SLAM. In both conventional CT-SLAM and the proposed method, motion distortion caused by the moving platform is effectively handled by deploying continuous-time trajectory representation, but the proposed method is able to prevent the global optimization cost from increasing according to the operation time and is therefore more amenable for real-time applications.

[1] https://youtu.be/QNNLncT9XmQ

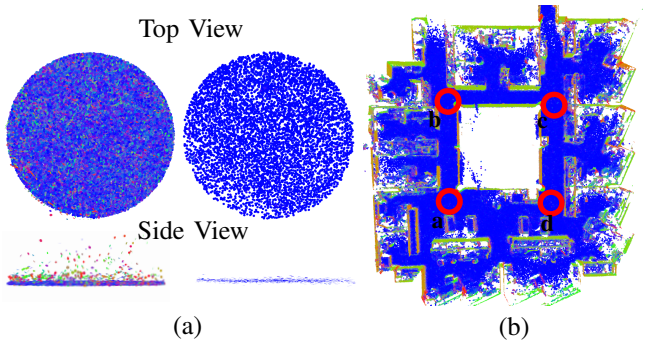

Fig. 8: (a) Qualitative comparison on a well-known structure. A circular patch is extracted from the floor of the map in (b) and compared with the patch from [3]. Top and side view of the patch from CT-SLAM [3] (left) and our method (right). (b) Extracted patch locations in Table III. Each patch has 0.7m radius.

| Location | No. Points | No. Surfel | CT-SLAM [3] | Proposed |
|---|---|---|---|---|
| a | $47.8 \times 10^4$ | $3.7 \times 10^3$ | 16.08 | **7.72** |
| b | $37.8 \times 10^4$ | $4.1 \times 10^3$ | 15.78 | **5.79** |
| c | $40.6 \times 10^4$ | $3.8 \times 10^3$ | 16.43 | **10.39** |
| d | $56.3 \times 10^4$ | $3.8 \times 10^3$ | 19.40 | **13.07** |

TABLE III: Surface estimation statistics. The projective distance errors in the last two columns are in mm. The maximum ideal number of 0.02m surfels that can fit into 0.7m circular patch is $3.8 \times 10^3$.

ACKNOWLEDGMENTS

The authors gratefully acknowledge funding of the project by the CSIRO and QUT. The institutional support of CSIRO and QUT, and the help of several individual members of staff in the Autonomous Systems Laboratory including Tom Lowe, Gavin Catt and Mark Cox are greatly appreciated.